\documentclass[10pt,twocolumn,letterpaper]{article}

\usepackage{cvpr}              
\definecolor{cvprblue}{rgb}{0.21,0.49,0.74}
\usepackage[pagebackref,breaklinks,colorlinks,allcolors=cvprblue]{hyperref}

\usepackage[table]{xcolor}
\usepackage{booktabs}
\usepackage{array}
\usepackage{adjustbox}
\usepackage{geometry}
\definecolor{myblue}{RGB}{200,230,255}  
\definecolor{lightgray}{gray}{0.9}       
\usepackage{makecell}

\def\paperID{} 
\def\confName{CVPR}
\def\confYear{2026}

\title{Efficient Hybrid SE(3)-Equivariant Visuomotor Flow Policy via Spherical Harmonics for Robot Manipulation}

\author{
    Qinglun Zhang$^{1,2}$\thanks{This work was done during an internship at Dexmal.} \quad
    Shen Cheng$^{2}$ \quad
    Tian Dan$^{1}$ \quad
    Haoqiang Fan$^{2}$ \quad  \\
    Guanghui Liu$^{1}$ \quad
    Shuaicheng Liu$^{1,2}$\thanks{Corresponding author.} \\
    $^1$University of Electronic Science and Technology of China \quad $^2$Dexmal \\
    {\tt\small \{zhangqinglun26@std., liushuaicheng@\}uestc.edu.cn},
    {\tt\small chengshen@dexmal.com}
}

\begin{document}
\maketitle
\begin{abstract}
While existing equivariant methods enhance data efficiency, they suffer from high computational intensity, reliance on single-modality inputs, and instability when combined with fast-sampling methods. In this work, we propose \textbf{E3Flow}, a novel framework that addresses the critical limitations of equivariant diffusion policies.  E3Flow overcomes these challenges, successfully unifying efficient rectified flow with stable, multi-modal equivariant learning for the first time. Our framework is built upon spherical harmonic representations to ensure rigorous SO(3) equivariance. We introduce a novel invariant Feature Enhancement Module (FEM) that dynamically fuses hybrid visual modalities (point clouds and images), injecting rich visual cues into the spherical harmonic features. We evaluate E3Flow on 8 manipulation tasks from the MimicGen and further conduct 4 real-world experiments to validate its effectiveness in physical environments. Simulation results show that E3Flow achieves a 3.12\% improvement in average success rate over the state-of-the-art Spherical Diffusion Policy (SDP) while simultaneously delivering a 7$\times$ inference speedup. E3Flow thus demonstrates a new and highly effective trade-off between performance, efficiency, and data efficiency for robotic policy learning.  Code: \url{https://github.com/zql-kk/E3Flow}.


\end{abstract}

\begin{figure}[t]
    \centering
    \includegraphics[width=0.95\linewidth]{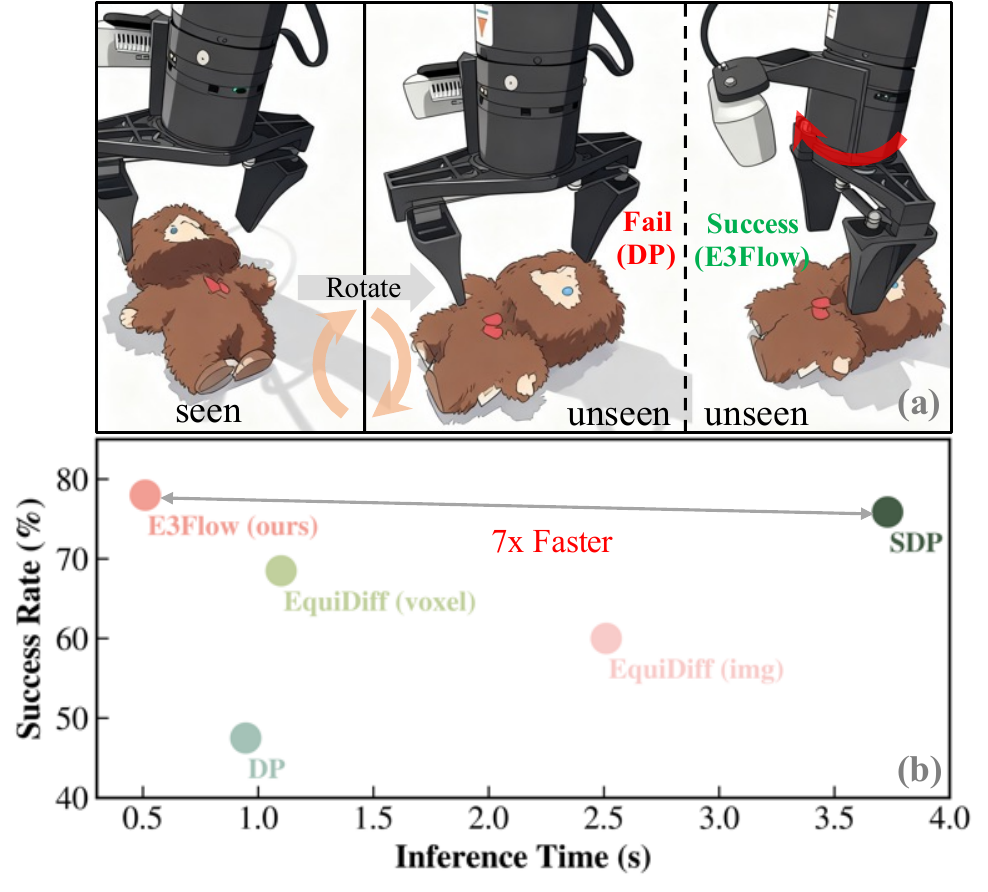}
    \vspace{-10pt}
    \caption{(a) Diagram of the E3Flow equivariant policy. E3Flow can learn equivariant trajectories under unseen scene transformations, whereas DP fails due to the lack of symmetry priors. (b) Comparison of average success rates and inference efficiency between state-of-the-art equivariant and non-equivariant policies on MimicGen tasks.}\vspace{-6pt}
    \label{teaser}
\end{figure}    
\section{Introduction}
\label{sec:intro}

Diffusion-based policies \cite{chi2025diffusion, zhu2025se, xu2026hero, xu2026actiongeometryprediction3dgeometric} have significantly advanced robotic policy learning, adeptly modeling complex, multimodal action distributions through behavior cloning \cite{mandlekar2021matters, florence2022implicit}. However, their reliance on large, high-quality expert datasets, which are costly and labor-intensive to acquire, hinders their real-world practicality. Consequently, enhancing data efficiency remains a critical challenge for robotic agents.

A highly effective strategy for improving data efficiency and generalization is leveraging symmetries inherent in robotic tasks. Since the physical world is replete with symmetries, this property, known as equivariant learning \cite{cohen2016group, gerken2023geometric, he2021efficient}, can drastically reduce data requirements. For instance, as shown in Fig.~\ref{teaser} (a), when the toy on the table is rotated to an unseen pose, the non-equivariant network (DP) \cite{chi2025diffusion} fails to grasp it due to the absence of symmetry priors, whereas our equivariant network (E3Flow) can successfully derive the correct grasping trajectory by correspondingly rotating the original one, without requiring any additional expert demonstrations. Extensive studies confirm that embedding such symmetry priors into diffusion models significantly boosts performance and data efficiency \cite{yang2024equibot, wang2024equivariant, zhu2025se, tie2024seed, ryu2024diffusion}.

Despite their promise, existing equivariant diffusion policies face two major hurdles. First, they are computationally intensive \cite{liao2022equiformer, liao2023equiformerv2}, limiting real-time use. Second, they typically rely on single-modality visual inputs (e.g., point clouds or images), which often miss the fine-grained details crucial for complex manipulation. These challenges restrict the potential of equivariant diffusion policies in real-world applications.

While fast sampling methods, such as one-step diffusion or flow-based models, have been proposed, they have not been successfully integrated with equivariant policies \cite{he2024consistency, frans2024one, yang2024consistencyflowmatchingdefining, geng2025meanflowsonestepgenerative}. Current imitation learning–based fast-sampling methods like FlowPolicy~\cite{zhang2025flowpolicy} and Consistency Policy~\cite{prasad2024consistencypolicyacceleratedvisuomotor} generally ignore symmetry priors. Meanwhile, as we demonstrate in Sec.~\ref{sec:Ablation}, directly applying these methods to equivariant policy learning leads to instability and performance degradation, particularly as task complexity increases. This leaves a critical gap: a method that combines the data efficiency of equivariance with the speed of fast sampling. 

In this work, we address these limitations by proposing E3Flow, a unified framework for multi-modality equivariant policy learning that is also highly sample efficient. Firstly, our method employs spherical harmonic representations \cite{liao2023equiformerv2} to achieve stable and rigorous SO(3)-equivariant embeddings, ensuring feature consistency under rotational transformations. Next, we integrate point clouds and images as two complementary visual modalities to enable fine-grained perception of complex scenes. To fuse these modalities, we further propose a feature enhancement module (FEM) that dynamically adjusts the weights between modalities, injecting fine-grained semantic cues from images into the spherical harmonic representations of point clouds, thereby significantly enhancing the expressiveness and visual sensitivity of the equivariant features.


We evaluate E3Flow on 8 manipulation tasks from the MimicGen benchmark, as illustrated in Fig.~\ref{teaser} (b). Our method achieves a 3.12\% improvement in average success rate over the strongest baseline, Spherical Diffusion Policy (SDP) \cite{zhu2025se}, while simultaneously delivering a 7$\times$ speedup in inference. Real-robot experiments further validate E3Flow’s substantial progress toward practical deployment of equivariant policies.

In summary, our key contributions are as follows:
\begin{itemize}
    \item  We present E3Flow, the first flow matching policy based on spherical harmonic representations, which generates high-quality actions through stable SO(3) embeddings and efficient flow modeling.
    \item  We design an invariant feature enhancement module (FEM) under hybrid visual modalities, dynamically injecting fine-grained visual cues into 3D invariant features for more expressive representations.
    \item  We conduct extensive experiments on both the MimicGen benchmark and in the real-world to demonstrate the effectiveness of E3Flow.
\end{itemize}

\section{Related Work}
\label{sec:realted work}

\begin{figure*}[t]
    \centering
    \includegraphics[width=\linewidth]{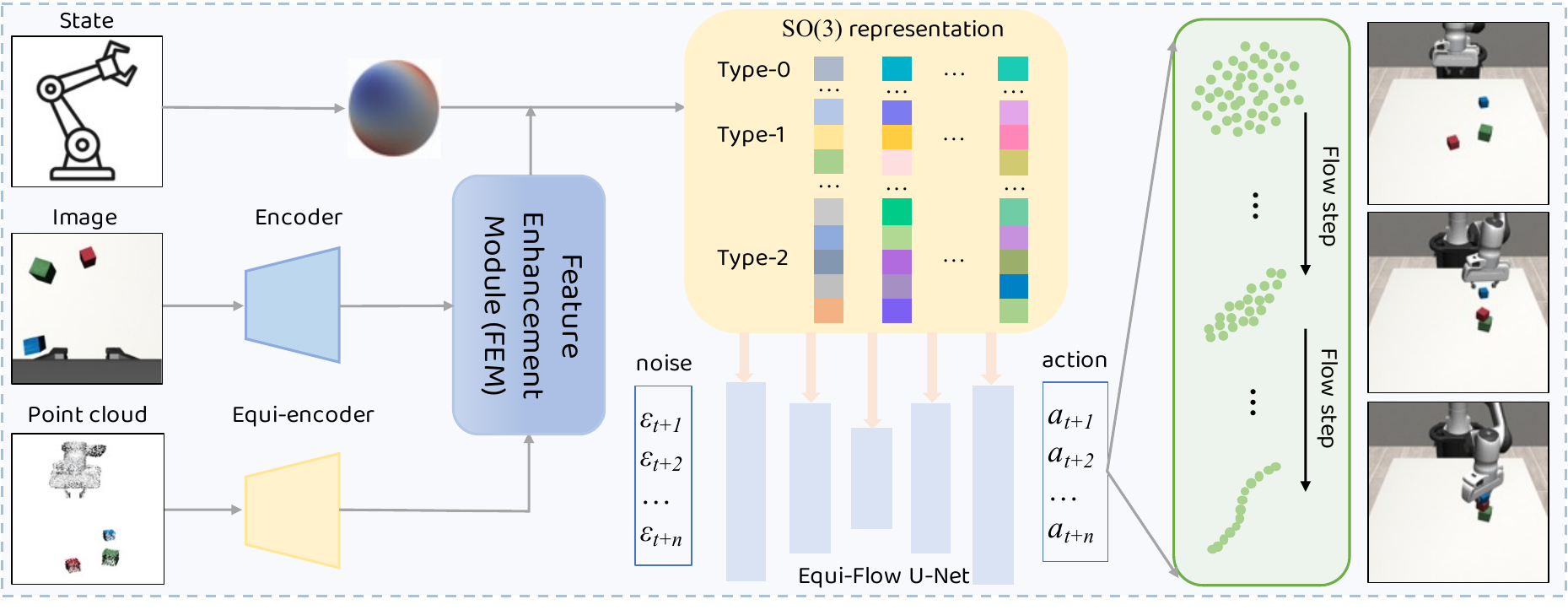}
    \caption{Overall pipeline. E3Flow encodes multimodal inputs through equivariant and non-equivariant visual encoders, aligns invariant visual features across modalities, and constructs a spherical harmonic–equivariant representation to efficiently guide flow matching for generating high-quality equivariant actions.}
    \label{pipeline}
\end{figure*}

\subsection{Imitation Learning with Diffusion Models}

Imitation learning (IL) \cite{hussein2017imitation}, particularly through behavioral cloning (BC), faces two primary challenges: cumulative errors \cite{zhao2023learning} and limited generalization due to the multimodality of expert demonstrations. Action chunking can address the former \cite{zhao2023learning}, diffusion and flow-based models are increasingly used to handle the latter, proving effective for modeling multimodal actions in robotics \cite{chi2025diffusion, zhang2025flowpolicy, zhang2025reinflow}. However, the performance of these generative models is fundamentally constrained by the expert dataset. Specifically, data quality determines the fidelity of the generated actions, while data diversity dictates the model's generalization ability. Therefore, enhancing data efficiency emerges as a key factor in improving an agent's generalization.

\subsection{Equivariant Learning in Robotics}
Enhancing data inputs and refining architectures boosts data efficiency \cite{luo2026DMAligner}. Prior work has leveraged richer 3D information \cite{ze20243d, wang2024rise, du2025hierarchical}, powerful foundation models \cite{kim2024openvla, black2024pi_0, jia2025lift3d}, and hybrid visual 3D representations \cite{zhong2025freqpolicy}. Alternatively, learning the abundant 3D symmetries in the physical world can significantly boost efficiency and generalization to unseen object poses. In robotics, equivariant learning has been applied to pose estimation \cite{wan2025equivariant, du2025se}, grasp detection \cite{zhu2023robot, lim2024equigraspflow}, reinforcement learning (RL) \cite{wang2022so2, zhao2024equivariant}, and imitation learning \cite{yang2024equibot, wang2024equivariant, yang2024equivact, zhu2025se}. However, most equivariant policies still rely on supervised or RL frameworks, requiring large-scale interaction data and heavy computation. By contrast, combining equivariant representations with generative modeling offers an efficient, symmetry-consistent approach. Our work integrates SE(3)-equivariant modeling and multi-modal enhanced representations with flow-based policy learning to achieve both data efficiency and fast visuomotor inference.

\subsection{Equivariant Diffusion Policy Learning}

Recent generative models have become popular in robotics for their strong multimodal action generation capabilities. One critical line of research has focused on equivariant learning to ensure policies generalize across different 3D poses. This has led to several diffusion-based methods, including EquiBot \cite{yang2024equibot} (using vector neurons), EquiDiff \cite{wang2024equivariant} (discrete SO(2) convolutions), ET-SEED \cite{tie2024seed} (SE(3)-Transformers), and SDP \cite{zhu2025se} (spherical harmonics). However, these methods inherit a major drawback of diffusion models \cite{ho2020denoising}: they rely on hundreds of iterative denoising steps, which significantly increases the inference latency of already complex equivariant networks. In addition, they suffer from other limitations, such as dependence on precise segmentation (EquiBot, ET-SEED), discretization errors (EquiDiff), or vulnerability to occlusions (SDP). Another line of research focuses on accelerating generative models, where flow matching \cite{lipman2022flow, hu2024adaflow, klein2023equivariant, RAW-Flow} enables efficient sample generation with only a few integration steps by directly learning a continuous flow field.


Despite its potential for speed, applying flow matching to create an equivariant policy is non-trivial. Most equivariant models have focused on the diffusion framework, and adapting these complex geometric constraints to a direct flow-matching formulation remains a significant challenge. To address these gaps, we introduce a simpler yet more efficient flow-based formulation integrated with equivariant networks.

\section{Method}
\label{sec:method}

\subsection{Preliminaries}

\paragraph{Equivariance} characterizes a property of transformation consistency. Intuitively, when the input undergoes a certain transformation, the model’s output transforms in a corresponding manner. 
Formally, let \( f: \mathcal{X} \rightarrow \mathcal{Y} \) be a mapping, and let \( G \) be a group acting on the input space \( \mathcal{X} \). 
If there exists a corresponding transformation \( \rho(g) \) acting on the output space such that
\begin{equation}
    f(\rho(g) \, x) = {\rho(g)} \, f(x), \quad 
    \forall g \in G, \, x \in \mathcal{X},
\end{equation}
then the function \( f \) is said to be equivariant with respect to the group \( G \).

The group \( SO(2) \) denotes the rotation group in 2D Euclidean space, 
which describes all possible rotations of the 2D plane around the origin. 
Similarly, \( SO(3) \) represents the rotation group in 3D space, 
encompassing all rotations about arbitrary axes by arbitrary angles. 
The group \( SE(3) \), known as the special Euclidean group, 
is composed of both 3D rotations and translations, 
and thus represents all rigid-body transformations in 3D space.
The trivial representation $\rho_0$ is a scalar representation in which all group elements act as the identity transformation. 
An irreducible representation (irrep) refers to a representation that cannot be decomposed into smaller invariant subspaces; 
it serves as the fundamental building block of group representations and is typically expressed as a vector $\rho_1$ or a higher-order tensor $\rho_2$. 
Therefore, a general group action $\rho$ can be expressed as a direct sum of the trivial and irreducible representations:
\begin{equation}
    \rho = \rho_0 \oplus \rho_1 \oplus \rho_2
\end{equation}

\paragraph{Spherical Harmonics} denoted as $Y_l^m(\theta, \phi)$, are orthogonal basis functions defined on the unit sphere with angular coordinates $(\theta, \phi)$. Here, $l \ge 0$ indicates the degree and $m \in [-l, l]$ denotes the order. Any function $f(\theta, \phi)$ defined on the sphere can be expanded as a linear combination of these basis functions:
\begin{equation}
f(\theta, \phi) = \sum_{l=0}^{\infty} \sum_{m=-l}^{l} c_l^m Y_l^m(\theta, \phi)
\end{equation}
where $c_l^m$ are the corresponding expansion coefficients. Under a rotation $R \in \mathrm{SO(3)}$, the degree $l$ remains unchanged, while the spherical harmonics of the same degree undergo a linear mixing:
\begin{equation}
Y_l^m(R^{-1} \hat{\mathbf{r}}) = \sum_{m'=-l}^{l} D_{m m'}^{(l)}(R) \, Y_l^{m'}(\hat{\mathbf{r}})
\end{equation}
where $D^{(l)}(R)$ is the $l$-th order irreducible representation matrix of the rotation group $\mathrm{SO(3)}$. By representing inputs in the spherical harmonics, the network can produce predictable equivariant outputs under rotations.


\begin{figure}[t]
    \centering
    \includegraphics[width=\linewidth]{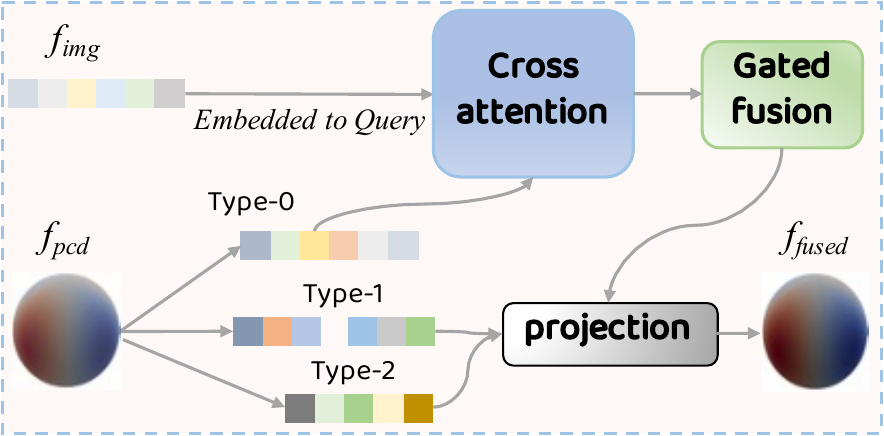}
    \caption{Illustration of the Feature Enhancement Module (FEM). FEM injects semantic information from images into the equivariant representation of point clouds, achieving efficient fusion of semantic and geometric features.}
    \label{fig3}
\end{figure}

\begin{figure*}[t]
    \centering
    \includegraphics[width=\linewidth]{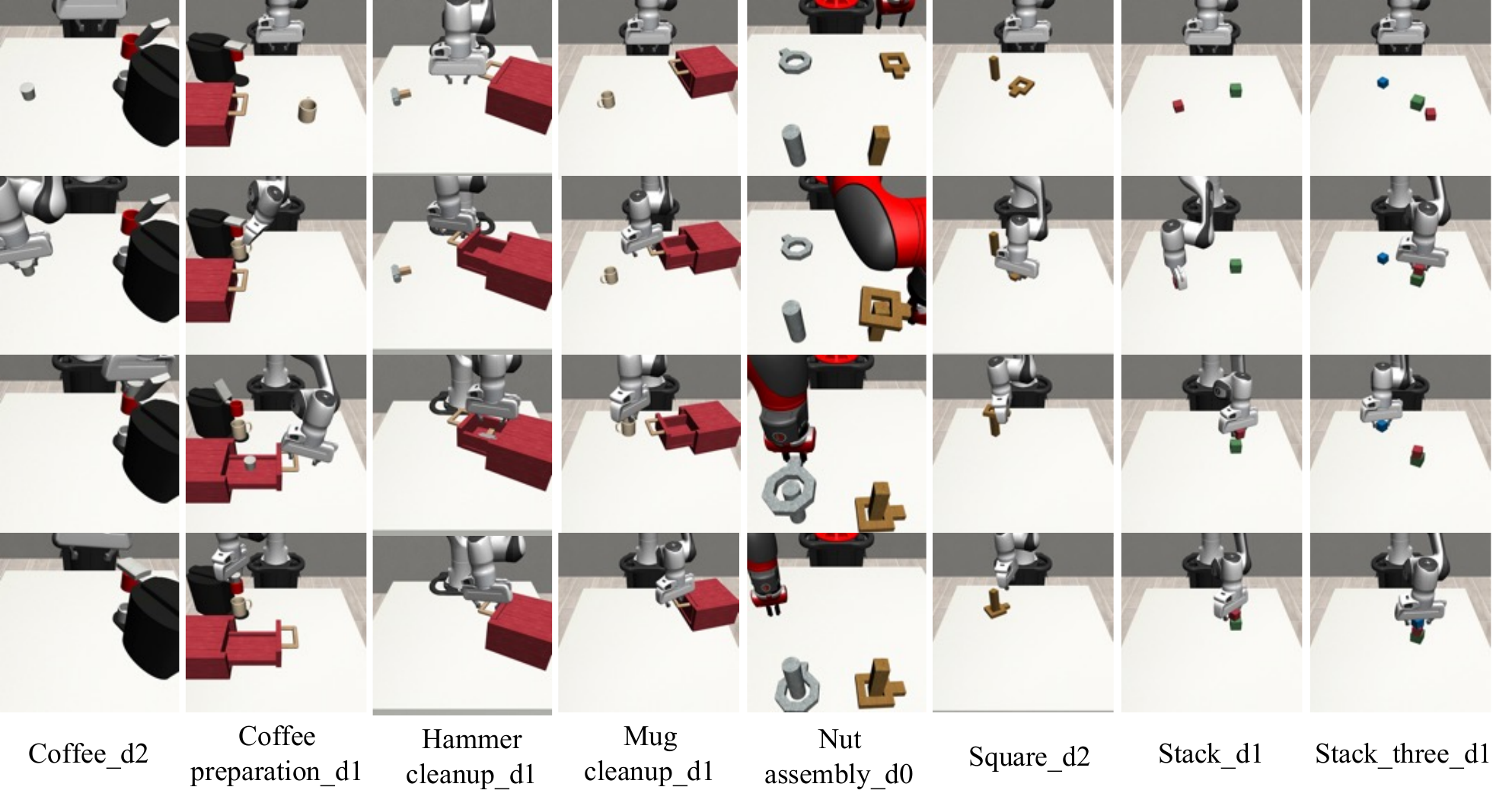}
    \caption{Visualization of the execution process of E3Flow across eight tasks from MimicGen. Each column depicts the task progression, where the top corresponds to the task initialization and the bottom indicates the final completion state.}
    \label{fig:placeholder}
    \label{fig4}
\end{figure*}


\subsection{Overview}

Our goal is to enable an intelligent agent to learn robust and generalizable policies $\pi$ from a limited number of expert demonstrations. The inherent symmetries in the learning environment can significantly improve both data efficiency and generalization capability. Therefore, we aim to incorporate symmetry learning into the policy network, allowing it to learn an equivariant mapping from observations $O(o_1, o_2, \ldots ,o_n)$ to actions $A(a_1, a_2, \ldots ,a_n)$. Since both observations $O(o_1, o_2, \ldots ,o_n)$ and actions $A(a_1, a_2, \ldots ,a_n)$ are equivariant under the same transformation group, the optimal policy $\pi((o_1, a_1), (o_2,a_2), \ldots, (o_n, a_n))$ should also be equivariant.
To better representation equivariant features and guide efficient action generation, we propose E3Flow, a hybrid equivariant flow policy based on spherical representations. As illustrated in Fig.~\ref{pipeline}, the E3Flow takes hand-eye images and single-view point clouds as visual observations. These inputs are processed by 2D and 3D encoders to extract both equivariant and invariant features. A feature enhancement module (FEM) is then applied to strengthen invariant representations. The resulting features, combined with proprioceptive embeddings projected onto the sphere, serve as the conditional representation for the flow model. Finally, E3Flow generates equivariant actions through fast and efficient flow matching.

\subsection{Spherical Harmonic Visual Representation}

This section introduces how we construct an SO(3)-equivariant observation encoder with feature enhancement. Our input consists of two parts: (1) visual information include hand-eye images and single-view point clouds, and (2) proprioceptive states of the robot. For the visual modality, we employ a ResNet to extract invariant features from images and an EquiformerV2 to extract equivariant features from point clouds. The invariant image features are injected into the point cloud features via a Feature Enhancement Module (FEM) to enhance semantic understanding. The overall workflow of the proposed visual representation is illustrated in Fig.~\ref{fig3}. FEM injects invariant image features into the scalar (Type-0) component via cross-modal attention ($\mathcal{A}$). A gating mechanism ($\Lambda$) adaptively balances the image contribution to form the final spherical representation:
\begin{equation}
f_\mathrm{fused} = \Pi\Big[\Lambda\big(\mathcal{A}(f_\text{pcd}^{(0)}, f_\mathrm{img}), f_\text{pcd}^{(0)}\big) \,\Vert\, f_\text{pcd}^{(>0)}\Big]
\end{equation}
where $\Pi(\cdot)$  denotes projection, \(f_\text{pcd}^{(0)}\) and \(f_\text{pcd}^{(>0)}\) are the scalar and higher-order spherical point cloud features, and \(f_\text{img}\) is the invariant image feature. The robot's proprioceptive state $s$ consists of the 3D position, a 6D rotation, and a 1D gripper state. Specifically, the 3D position and 6D rotation are treated as equivariant information and embedded into spherical features as vectors, while the gripper state is modeled as a trivial representation.


The proposed visual representation offers several notable advantages. First, representing features as continuous spherical Fourier coefficients naturally ensures SO(3) compatibility and effectively reduces EquiDiff’s discretization errors. Second, unlike EquiBot, EquiformerV2 encodes higher-order coefficients, enabling the capture of finer directional and rotational details. Third, compared with ET-SEED, the visual encoder requires only a single forward pass, greatly accelerating both training and inference. Finally, in contrast to SDP, which relies solely on sparse point clouds, our method leverages hybrid visual inputs together with FEM-injected invariant features to capture fine-grained spatial details, particularly in challenging pick-and-place tasks.

\subsection{Equivariant Action via Rectified Flow}

This section introduces how to construct an efficient equivariant action generation framework through rectified flow. The action shares the same geometric structure as the state $s$. We first revisit the rectified flow formulation for fast equivariant action generation, and then demonstrate how rectified flow can be naturally adapted to equivariant networks. Rectified flow aims to parameterize a vector field under two marginal constraints $p_{t_0}=0$ and $p_{t_1}=1$, such that the induced probability path $p_t$ continuously transports data from a source distribution ($x^{\text{src}} {\sim} \Gamma^{\text{src}}$) to the target distribution ($a^{\text{tar}} {\sim} \Gamma^{\text{tar}}({\cdot}|s,v)$), conditioning by state $s$ and visual representation $v$. This process is typically described by an ordinary differential equation (ODE):
\begin{equation}
     \left\{\begin{aligned}
     \frac{d{\xi}_{x}(t)}{dt} &={\nu}_{\theta}(t,{\xi}_x(t), s, v)\\[4pt]
     {\xi}_x(0) &=x
     \end{aligned}\right.
\end{equation}
${\xi}_x(t)$ is called a flow. The learning objective is to match the predicted velocity ${\nu}{\theta}(t,x_t)$ with the true flow velocity $u(t,x_t)$, formulated as:
\begin{equation}\label{eq:2}
    \mathcal{L}_{\text{RF}}(\theta)=E_{t,p_t}{\Vert{{\nu}_{\theta}(t,x_t, s, v)-u(t,x_t, s, v)}\Vert}^2_2
\end{equation}
Since the true velocity field $u(t,x_t)$ and the intermediate distribution $p_t$ are unknown, rectified flow constructs linear interpolations between samples from the source and target distributions:
\begin{equation}
    x_t = (1 - t)x_0 + t a
\end{equation}
The corresponding true velocity at time $t$ is:
\begin{equation}
    v_t^*(x_t) = a - x_0
\end{equation}
Accordingly, the training loss can be defined as:
\begin{equation}
    \mathcal{L}_{\text{RF}}(\theta) = 
    \mathbb{E}_{t, x_0, x_1}
    \left[
        \left\| 
        v_\theta(x_t, t, s, v) - (a - x_0)
        \right\|^2
    \right]
\end{equation}
The core idea of rectified flow is to solve a simple nonlinear least-squares optimization problem to learn an ODE that evolves as closely as possible along the 
straight paths connecting points sampled from the source distribution and the target distribution.
It requires no additional parameters beyond those used in standard flow matching. Finally, since the parameterization network of the velocity field is equivariant, we have:
\begin{equation}
    v_\theta(\rho*x_t, t, \rho*s, \rho*v)=\rho*v_\theta(x_t, t, s, v)
\end{equation}
At the same time, since the training objective is a linear transformation of equivariant actions, the loss remains invariant under group actions, and thus there exists an equivariant optimal solution. Therefore, our proposed E3Flow is end-to-end equivariant.

\section{Experiments}
\label{sec:Experiments}

\begin{table*}[tb]
\caption{Comparison of success rates between state-of-the-art equivariant and non-equivariant policies and the proposed method.
Eight representative tasks from the MimicGen benchmark are evaluated, each trained with the same 100 expert demonstrations under three random seeds. All models are tested under identical initialization conditions, and the maximum success rate is reported. Training and evaluation are performed on a single H20 GPU. Bold indicates the best result, and \underline{underline} denotes the second best.}
\label{table1}
\centering
\small  
\renewcommand{\arraystretch}{1.15}

\begin{adjustbox}{max width=\textwidth}
\begin{tabular}{
    p{2.7cm}  
    >{\centering\arraybackslash}p{1.2cm}
    >{\centering\arraybackslash}p{1.2cm}
    >{\centering\arraybackslash}p{1.2cm}
    >{\centering\arraybackslash}p{1.2cm}
    >{\centering\arraybackslash}p{1.2cm}
    >{\centering\arraybackslash}p{1.2cm}
    >{\centering\arraybackslash}p{1.2cm}
    >{\centering\arraybackslash}p{1.2cm}
    >{\centering\arraybackslash}p{1.2cm}
    >{\centering\arraybackslash}p{1.4cm}
}
\toprule
\textbf{Task}\rule{0pt}{3ex}&
\textbf{EquiBot} &
\textbf{BC-RNN} &
\textbf{ACT} &
\textbf{DP3} &
\textbf{DP} &
\textbf{EquiDiff (img)} &
\textbf{EquiDiff (voxel)} &
\textbf{SDP (DDiM)} &
\textbf{SDP (DDPM)} &
\textbf{E3Flow (ours)} \\
\midrule
Coffee\_D2             & 0  & 32 & 20 & 34 & 44 & 60 & \textbf{65} & 56 & 63 & \underline{64} \\
Coffee\_Preparation\_D1 & 2  & 10 & 36 & 8  & 64 & \underline{72} & \textbf{80} & 50 & 56 & 60 \\
Hammer\_Cleanup\_D1    & 14 & 26 & 32 & 56 & 48 & 54 & 70 & \underline{74} & \underline{74} & \textbf{84} \\
Mug\_Cleanup\_D1       & 24 & 20 & 26 & 22 & 46 & \underline{54} & 53 & 38 & \textbf{60} & \textbf{60} \\
Nut\_Assembly\_D0      & 3  & 31 & 41 & 12 & 54 & 72 & 67 & 88 & \underline{92} & \textbf{94} \\
Square\_D2             & 0  & 6  & 2  & 6  & 10 & 22 & 39 & 58 & \underline{64} & \textbf{70} \\
Stack\_D1              & 0  & 58 & 34 & 58 & 82 & 90 & 98 & \textbf{100} & \textbf{100} & \textbf{100} \\
Stack\_Three\_D1       & 0  & 8  & 4  & 8  & 32 & 56 & 76 & 94 & \underline{98} & \textbf{100} \\
\midrule
\rowcolor{lightgray}
\textbf{Average}           & 5.38 & 23.88 & 24.38 & 25.50 & 47.50 & 60.00 & 68.50 & 69.75 & \underline{75.88} & \textbf{79.00} \\
\bottomrule
\end{tabular}
\end{adjustbox}
\end{table*}

\subsection{Dataset and Implementation Details}

\paragraph{Simulation Benchmarks}

We validate E3Flow on the MimicGen simulator \cite{mandlekar2023mimicgen}, which offers realistic rendering and rich contact dynamics. Serving as a large-scale framework covering diverse tasks and robot embodiments, it provides a foundation for imitation learning and policy generalization. MimicGen includes long-horizon, high-precision assembly, and mobile manipulation tasks. Unlike script-based demonstrations, all expert trajectories come from real human demonstrations and are automatically expanded into diverse, multimodal trajectories, introducing strong non-Markovian characteristics that closely approximate real-world dynamics and making it a reliable benchmark for robot learning.


\paragraph{Baselines}
We aim to enhance the data efficiency of our model through hybrid spherical harmonic equivariant representations and introduce flow matching into equivariant action generation to further improve inference efficiency and action quality. Accordingly, we primarily compare against state-of-the-art equivariant and non-equivariant baselines for thorough evaluation. Equivariant baselines include SDP \cite{zhu2025se}, a diffusion-based policy architecture built upon spherical Fourier representations with $\mathrm{SE}(3)$-equivariance; EquiDiff \cite{wang2024equivariant}, a discrete $\mathrm{SO}(2)$-equivariant diffusion framework supporting 2D and 3D inputs; and EquiBot \cite{yang2024equibot}, a $\mathrm{SIM}(3)$-equivariant model based on vector neurons. Non-equivariant baselines include ACT \cite{zhao2023learning}, a transformer-based policy performing action chunking; DP \cite{chi2025diffusion}, a diffusion architecture with image inputs; DP3 \cite{ze20243d}, an efficient diffusion framework on single-view point cloud inputs; and BC-RNN \cite{mandlekar2021matters}, a behavior cloning approach leveraging Gaussian mixture models and recurrent neural networks to capture multimodal action distributions. These baselines cover a wide spectrum of architectural paradigms and input modalities, enabling a comprehensive and rigorous evaluation of our proposed E3Flow across both equivariant and non-equivariant policy learning settings.


\paragraph{Implementation Details}

We select 8 tasks from MimicGen as benchmarks, each containing 100 expert demonstrations and providing RGB-D observations from both the hand-eye and front cameras of the manipulator. For image observations, the resolution is cropped to $84 \times 84$. For point cloud observations, we apply farthest point sampling to downsample each point cloud to 1024 points while preserving color. Following DP3 preprocessing, we retain only the points within the workspace. For voxel observations, the voxel grid size is set to $84^3$. Experimental details and results in the real-world are provided in the supplementary material.

During training, we employ the AdamW optimizer with a learning rate of $1\times10^{-4}$ and a batch size of 64 to update the model parameters. Exponential Moving Average (EMA) is used to ensure training stability, with a decay rate of 0.95. All our models and baselines were implemented in PyTorch and trained on a single NVIDIA H20 GPU for 500 epochs. 
Evaluation was conducted every 20 epochs with 50 episodes per test, and we reported the maximum success rate achieved across the entire evaluation process.


\begin{table*}[tb]
\caption{Comparison of inference time (s) between state-of-the-art equivariant and non-equivariant policies and the proposed method. We evaluate eight tasks from MimicGen and compute average inference time across all steps. E3Flow achieves an inference speed 7× faster than the strongest baseline. Inference runs on a single H20 GPU, and we bold the top-3 results.}
\label{table2}

\centering
\footnotesize  
\renewcommand{\arraystretch}{1.15}
\begin{tabular}{
    p{2.7cm}  
    >{\centering\arraybackslash}p{1.2cm}
    >{\centering\arraybackslash}p{1.2cm}
    >{\centering\arraybackslash}p{1.2cm}
    >{\centering\arraybackslash}p{1.2cm}
    >{\centering\arraybackslash}p{1.2cm}
    >{\centering\arraybackslash}p{1.2cm}
    >{\centering\arraybackslash}p{1.2cm}
    >{\centering\arraybackslash}p{1.4cm}
}
\toprule
\textbf{Task}\rule{0pt}{3ex}&
\textbf{EquiBot} &
\textbf{DP3} &
\textbf{DP} &
\textbf{EquiDiff (img)} &
\textbf{EquiDiff (voxel)} &
\textbf{SDP (DDiM)} &
\textbf{SDP (DDPM)} &
\textbf{E3Flow (ours)} \\
\midrule
Coffee\_D2             & 2.06  & 0.103 & 0.95 & 2.44 & 1.10 & 0.49 & 3.79 & 0.51 \\
Coffee\_Preparation\_D1 & 2.13  & 0.112  & 0.89 & 2.07 & 1.11 & 0.44 & 3.46 & 0.51 \\
Hammer\_Cleanup\_D1    & 1.92 & 0.105 & 0.94 & 2.36 & 1.11 & 0.48 & 3.67 & 0.51 \\
Mug\_Cleanup\_D1       & 1.91 & 0.113 & 1.01 & 3.68 & 1.10 & 0.46 & 3.98 & 0.52 \\
Nut\_Assembly\_D0      & 1.80  & 0.102 & 0.94 & 2.34 & 1.09 & 0.45 & 3.71 & 0.49 \\
Square\_D2             & 2.07  & 0.112  & 0.95 & 2.39 & 1.11 & 0.46 & 3.85 & 0.54 \\
Stack\_D1              & 2.16 & 0.112 & 0.95 & 2.41 & 1.09 & 0.46 & 3.84 & 0.50 \\
Stack\_Three\_D1       & 2.17 & 0.111  & 0.94 & 2.38 & 1.10 & 0.45 & 3.56 & 0.53 \\
\midrule
\rowcolor{lightgray}
\textbf{Average}           & 2.03 & \textbf{0.109} & 0.95 & 2.51 & 1.10 & \textbf{0.46} & 3.73 & \textbf{0.51} \\
\bottomrule
\end{tabular}
\end{table*}

\subsection{Comparison with Various Baselines}

\paragraph{Quantitative Comparisons on Success Rate}
Table~\ref{table1} presents the success rates of each model across 8 tasks, while Fig~\ref{fig4} visualizes the execution process of E3Flow for each task. Except for EquiBot, all equivariant policies significantly outperform their non-equivariant counterparts, demonstrating the superior generalization and robustness of equivariant designs in robot policy learning under limited data. The failure of EquiBot stems from its handling of $\mathrm{SIM}(3)$ transformations, which involve scaling the entire scene. While effective for single-object manipulation, this operation disrupts point cloud structures in complex environments. Compared to the strongest non-equivariant 3D baseline, DP3, E3Flow achieves a 53.5\% higher average success rate. Furthermore, compared to the strongest equivariant 3D baseline, SDP, E3Flow still improves performance by 3.12\%, validating the effectiveness of refining equivariant action generation with invariant visual semantics under hybrid inputs.  Nevertheless, compared to EquiDiff (voxel), E3Flow attains a 10.5\% higher average success rate, showing that modeling continuous equivariance is more effective than discrete equivariance. Overall, E3Flow achieves the best average performance by injecting invariant features from hand-eye images into point cloud features, ensuring high-quality action generation.

\paragraph{Comparison Results of Inference Efficiency}
Table~\ref{table2} reports the average inference time per task and the mean inference time across all baselines. E3Flow achieves competitive results. The highest inference efficiency is observed in DP3, benefiting from its compact 3D representation. However, its simple MLP architecture struggles to capture directional information, leading to suboptimal success rates. Among all equivariant baselines, E3Flow achieves the fastest inference speed, being 7$\times$ faster than SDP (0.51s vs.\ 3.73s), 5$\times$ faster than EquiDiff (img) (0.51s vs.\ 2.51s), and 2$\times$ faster than EquiDiff (voxel) (0.51s vs.\ 1.10s). Furthermore, when replacing SDP's solver with the faster DDIM, the average performance dropped by 6.13\%, while E3Flow outperformed it by 8.25\% in success rate. These results demonstrate that in equivariant learning tasks, rectified flow defined by ordinary differential equations enables more efficient multimodal processing and generates higher quality actions with fewer inference steps.



\begin{table*}[tb]
\centering
\caption{Comparison of success rates between state-of-the-art policies and the proposed E3Flow under 0° and 10° perturbations. We train all models in tabletop-level scenes and directly perform zero-shot transfer to locally initialized SE(3) scenes, demonstrating the strong generalization ability of E3Flow. Bold indicates the best result, and \underline{underline} denotes the second best.}
\label{table3}
\footnotesize
\renewcommand{\arraystretch}{1.15}

\begin{adjustbox}{max width=\textwidth}
\begin{tabular}{
    p{2.5cm}  
    >{\centering\arraybackslash}p{1.2cm}  
    *{8}{>{\centering\arraybackslash}p{1.05cm}}  
}
\toprule
\textbf{Model} & \textbf{Equ.} &
\multicolumn{2}{c}{\textbf{Coffee\_D2}} &
\multicolumn{2}{c}{\textbf{Square\_D2}} &
\multicolumn{2}{c}{\textbf{Nut\_Assembly\_D0}} &
\multicolumn{2}{c}{\textbf{Hammer\_Cleanup\_D1}} \\
\cmidrule(lr){3-10}
 & & 0° & 10° & 0° & 10° & 0° & 10° & 0° & 10° \\
\midrule
EquiBot          & SIM(3) & 0 & 0 & 0 & 0 & 3 & 0 & 14 & 4 \\
DP               & N/A     & 44 & 6 & 10 & 0 & 54 & 3 & 48 & 18 \\
EquiDiff (voxel) & SO(2)  & \textbf{65} & 20 & 39 & 16 & 67 & 6 & 70 & 42 \\
SDP              & SE(3)  & 63 & \underline{32} & \underline{64} & \underline{29} & \underline{92} & \underline{39} & \underline{74} & \underline{53} \\
\rowcolor{gray!10}
E3Flow (ours)    & SE(3)  & \underline{64} & \textbf{38} & \textbf{70} & \textbf{34} & \textbf{94} & \textbf{52} & \textbf{84} & \textbf{58} \\
\bottomrule
\end{tabular}
\end{adjustbox}
\end{table*}

\paragraph{SE(3) Transformation Comparison}
Furthermore, we simulate local SE(3) variations by tilting both the table and the objects to rigorously evaluate the generalization ability of E3Flow. Specifically, all models are trained in standard tabletop-level scenes, while during testing, the table and objects are tilted by 10° along the y-axis for zero-shot evaluation. As shown in Table~\ref{table3}, both SDP and E3Flow adopt continuous SE(3) representations, making them more robust to SE(3) transformations and therefore superior to other models. Compared with SDP, E3Flow effectively alleviates the point cloud occlusion problem in SDP by leveraging fine-grained visual information. As a result, E3Flow achieves the best overall performance in unseen tilted scenes, demonstrating its strong generalization capability. Additionally, due to gravitational effects, we do not analyze scenarios with larger tilt angles.

\subsection{Ablation Study}
\label{sec:Ablation}

\begin{table}[htbp]
\caption{Ablation study evaluating the impact of input modality, fusion strategy, and generative method. “PCD”: point clouds, “Img”: hand-eye image, “RF”: rectified flow.}
\label{table4}

\centering
\small
\begin{adjustbox}{max width=0.48\textwidth}
\begin{tabular}{
    >{\centering\arraybackslash}p{1.8cm}
    >{\centering\arraybackslash}p{1.4cm}
    >{\centering\arraybackslash}p{1.8cm}
    >{\centering\arraybackslash}p{2.4cm}
}
\toprule
 \textbf{Input} & \textbf{Fusion} & \textbf{Generative} & \textbf{Success Rate (\%)} \\
\midrule
 PCD & - & RF & 75.88 \\
 PCD & - & Diffusion & 75.23 \\
 PCD+Img & \textit{cat} & RF & 72.36 \\
 PCD+Img & \textit{cat} & Diffusion & 71.42 \\
 PCD+Img & FEM & RF & \textbf{79.00} \\
 PCD+Img & FEM & Diffusion & \underline{77.58} \\
\bottomrule
\end{tabular}
\end{adjustbox}
\end{table}

\paragraph{Component Analysis of E3Flow}
We first conduct ablation studies on E3Flow to validate the effectiveness of each component, including the visual input modality, fusion strategy, and training paradigm. Table~\ref{table4} reports the average success rates across eight MimicGen tasks under different configurations. The results show that with single-modality visual input, the rectified flow approach achieves faster inference without sacrificing action quality. When adopting multimodal inputs, directly concatenating visual features from different modalities degrades performance due to the lack of alignment. Introducing the Feature Enhancement Module (FEM) proves essential, as it dynamically injects and gates fine-grained cues to adaptively select the most informative features for action generation, benefiting both training paradigms. Overall, these results confirm the effectiveness of E3Flow and highlight the importance of multimodal alignment in equivariant policy learning.

\begin{table}[htbp]
\caption{The effect of different flow matching training and sampling methods on success rate.}
\label{table5}

\centering
\small  
\begin{adjustbox}{max width=0.48\textwidth}
\begin{tabular}{lccccc}
\toprule
  & \textbf{MeanFlow} &  \textbf{$\alpha$Flow} &\textbf{RF-1} & \textbf{RF-5} &\textbf{RF-10} \\
\midrule
 Steps & 1 & 1 & 1 & 5 & 10 \\
 Inference Time (s)      & 0.17 & 0.17 & 0.16 & 0.28 & 0.51 \\
 Coffee\_D2              & 32  & 38   & 52   & \underline{56} & \textbf{64} \\
 Coffee\_Preparation\_D1 & 22  & 30   & 48   & \underline{52} & \textbf{60} \\
 Hammer\_Cleanup\_D1     & 70  & 68   & 76   & \underline{78} &  \textbf{84} \\
 Mug\_Cleanup\_D1        & 50  & \underline{50}   & 34   & 40 &  \textbf{60} \\
 Nut\_Assembly\_D0       & 74  & \underline{88}   & 82   & 80 &  \textbf{94} \\
 Square\_D2              & 26  & 58   & \underline{72}   & 68 & \textbf{70} \\
 Stack\_D1               & \underline{98}  & \textbf{100}   & \underline{98}  & \textbf{100} & \textbf{100} \\
 Stack\_Three\_D1        & 64  & 85   & 90  & \underline{94} & \textbf{100} \\
     Average             & 54.50 & 64.62 & 69.00 & \underline{71.00} & \textbf{79.00} \\
\bottomrule

\end{tabular}
\end{adjustbox}
\end{table}

\paragraph{Why Rectified Flow}
We further compare several efficient variants of flow matching, including MeanFlow \cite{geng2025meanflowsonestepgenerative}, AlphaFlow \cite{zhang2025alphaflowunderstandingimprovingmeanflow}, and Rectified flow, all of which are designed for one step or few step generation. Table~\ref{table5} reports the success rates across eight tasks under different training and sampling steps. The results show that simple tasks such as Stack\_D1 can produce high quality actions within a single step, but as task complexity increases, one-step sampling struggles to generate fine-grained actions, leading to lower success rates. Therefore, existing one-step sampling methods perform suboptimally when applied to equivariant models, as a single forward pass is insufficient for highly abstract equivariant features to effectively guide action generation. To address this, we adopt a balanced strategy by moderately increasing the number of inference steps to achieve a better trade-off between efficiency and performance.

\begin{figure}[t]
    \centering
    \includegraphics[width=\linewidth]{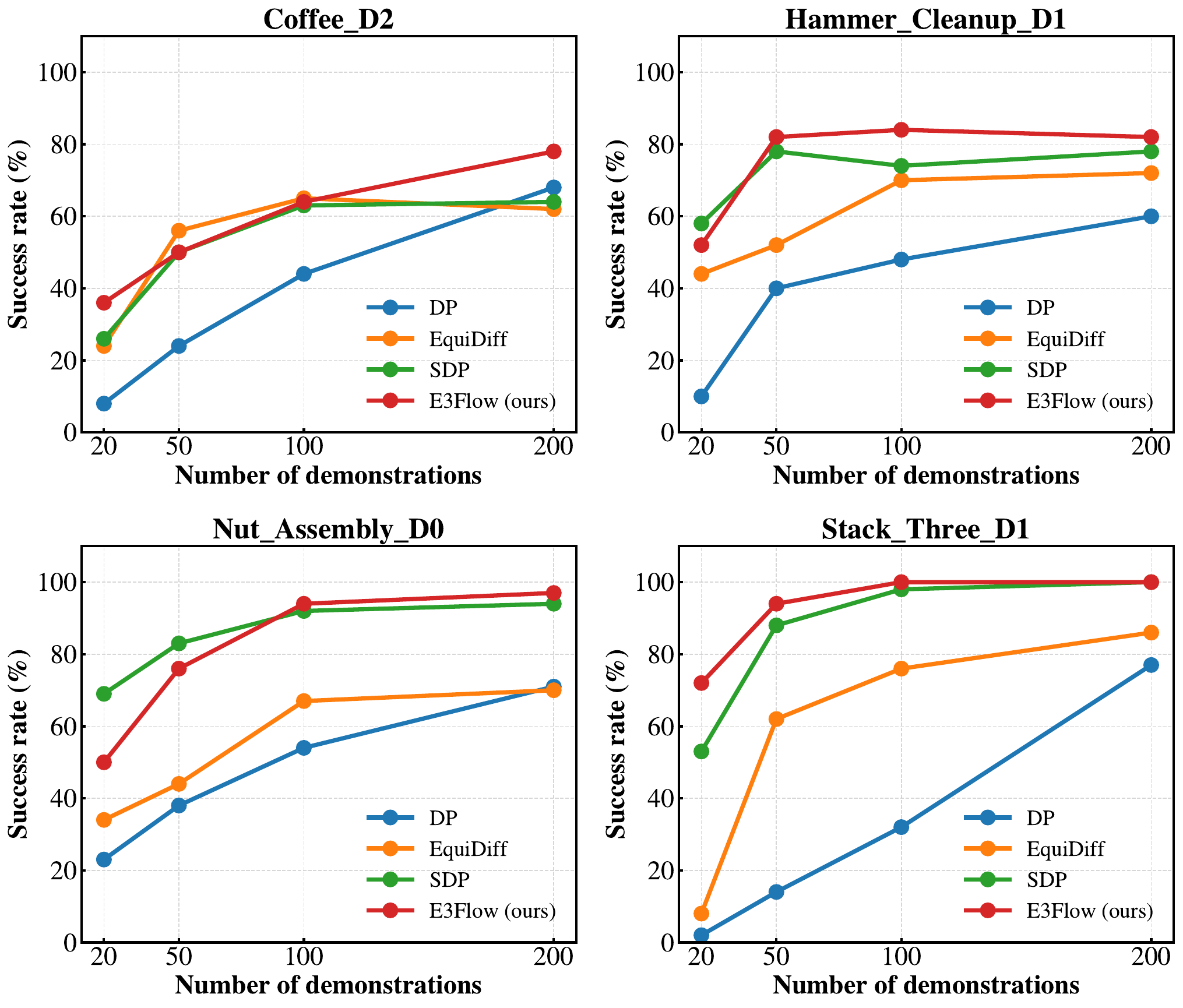}
    \caption{Ablation on the number of expert demonstrations. As the number of demonstrations increases, all methods show improved performance, while E3Flow consistently achieves higher success rates, highlighting its superior data efficiency.}
    \label{fig5}
\end{figure}

\paragraph{Data Efficiency}
Furthermore, we evaluate the data efficiency of E3Flow using four representative tasks and compare it with three strong baselines (DP, EquiDiff, and SDP) under varying numbers of expert demonstrations. As shown in Fig.~\ref{fig5}, success rates for all models increase with more demonstrations, confirming the dependence of imitation-based policies on demonstration quantity. Notably, E3Flow consistently outperforms other methods across data scales, achieving comparable performance to models trained with 200 demonstrations using only 100. These results highlight the superior data efficiency of our approach. Overall, integrating SE(3)-equivariant flow matching with fine-grained multimodal visual guidance enables robust generalization from limited data while producing more accurate, symmetry consistent actions, making E3Flow both data-efficient and practical for real-world robotic manipulation.

\section{Conclusion}
\label{sec:Conclusion}
We present E3Flow, an efficient hybrid SE(3)-equivariant visuomotor policy that seamlessly combines spherical harmonic representations with rectified flow to achieve fast, accurate, and symmetry-consistent robot control. Extensive experiments on MimicGen, as well as real-robot evaluations in realistic physical environments, demonstrate that embedding continuous symmetry priors into multimodal policy networks can simultaneously enhance generalization, inference efficiency, and action quality. In future work, we will systematically investigate the impact of both invariant and equivariant visual features on precise action generation. Additionally, exploring strict local equivariance is expected to further improve the data efficiency and robustness of equivariant policies in complex real-world scenarios.


{
    \small
    \bibliographystyle{ieeenat_fullname}
    \bibliography{main}

@String(CVPR= {IEEE Conf. Comput. Vis. Pattern Recog.})

@String(ICLR = {Int. Conf. Learn. Represent.})

@String(AAAI = {AAAI})

@String(CVPR  = {CVPR})

@String(ICLR  = {ICLR})

@article{hussein2017imitation,
  title={Imitation learning: A survey of learning methods},
  author={Hussein, Ahmed and Gaber, Mohamed Medhat and Elyan, Eyad and Jayne, Chrisina},
  journal={ACM Computing Surveys (CSUR)},
  volume={50},
  number={2},
  pages={1--35},
  year={2017},
  publisher={ACM New York, NY, USA}
}

@inproceedings{zhao2023learning,
  title={Learning fine-grained bimanual manipulation with low-cost hardware},
  author={Zhao, Tony Z and Kumar, Vikash and Levine, Sergey and Finn, Chelsea},
  booktitle={Proceedings of Robotics: Science and Systems (RSS)},
  year={2023}
}

@article{chi2025diffusion,
  title={Diffusion policy: Visuomotor policy learning via action diffusion},
  author={Chi, Cheng and Xu, Zhenjia and Feng, Siyuan and Cousineau, Eric and Du, Yilun and Burchfiel, Benjamin and Tedrake, Russ and Song, Shuran},
  journal={The International Journal of Robotics Research},
  volume={44},
  number={10-11},
  pages={1684--1704},
  year={2025},
  publisher={Sage Publications Sage UK: London, England}
}

@inproceedings{cohen2016group,
  title={Group equivariant convolutional networks},
  author={Cohen, Taco and Welling, Max},
  booktitle={International conference on machine learning (ICML)},
  pages={2990--2999},
  year={2016},
  organization={PMLR}
}

@article{gerken2023geometric,
  title={Geometric deep learning and equivariant neural networks},
  author={Gerken, Jan E and Aronsson, Jimmy and Carlsson, Oscar and Linander, Hampus and Ohlsson, Fredrik and Petersson, Christoffer and Persson, Daniel},
  journal={Artificial Intelligence Review},
  volume={56},
  number={12},
  pages={14605--14662},
  year={2023},
  publisher={Springer}
}

@inproceedings{he2021efficient,
  title={Efficient equivariant network},
  author={He, Lingshen and Chen, Yuxuan and Dong, Yiming and Wang, Yisen and Lin, Zhouchen and others},
  booktitle={Advances in Neural Information Processing Systems (NeurIPS)},
  volume={34},
  pages={5290--5302},
  year={2021}
}

@inproceedings{florence2022implicit,
  title={Implicit behavioral cloning},
  author={Florence, Pete and Lynch, Corey and Zeng, Andy and Ramirez, Oscar A and Wahid, Ayzaan and Downs, Laura and Wong, Adrian and Lee, Johnny and Mordatch, Igor and Tompson, Jonathan},
  booktitle={Conference on robot learning (CoRL)},
  pages={158--168},
  year={2022},
  organization={PMLR}
}

@inproceedings{zhang2025flowpolicy,
  title={Flowpolicy: Enabling fast and robust 3d flow-based policy via consistency flow matching for robot manipulation},
  author={Zhang, Qinglun and Liu, Zhen and Fan, Haoqiang and Liu, Guanghui and Zeng, Bing and Liu, Shuaicheng},
  booktitle={Proceedings of the AAAI Conference on Artificial Intelligence (AAAI)},
  volume={39},
  number={14},
  pages={14754--14762},
  year={2025}
}

@inproceedings{zhang2025reinflow,
  title={ReinFlow: Fine-tuning flow matching policy with online reinforcement learning},
  author={Zhang, Tonghe and Yu, Chao and Su, Sichang and Wang, Yu},
  booktitle={Advances in Neural Information Processing Systems (NeurIPS)},
  year={2025}
}

@inproceedings{ze20243d,
    title={3D Diffusion Policy: Generalizable Visuomotor Policy Learning via Simple 3D Representations},
    author={Yanjie Ze and Gu Zhang and Kangning Zhang and Chenyuan Hu and Muhan Wang and Huazhe Xu},
    booktitle={Proceedings of Robotics: Science and Systems (RSS)},
    year={2024}
}

@inproceedings{wang2024rise,
  title={Rise: 3d perception makes real-world robot imitation simple and effective},
  author={Wang, Chenxi and Fang, Hongjie and Fang, Hao-Shu and Lu, Cewu},
  booktitle={2024 IEEE/RSJ International Conference on Intelligent Robots and Systems (IROS)},
  pages={2870--2877},
  year={2024},
  organization={IEEE}
}

@inproceedings{kim2024openvla,
  title={Openvla: An open-source vision-language-action model},
  author={Kim, Moo Jin and Pertsch, Karl and Karamcheti, Siddharth and Xiao, Ted and Balakrishna, Ashwin and Nair, Suraj and Rafailov, Rafael and Foster, Ethan and Lam, Grace and Sanketi, Pannag and others},
  booktitle={Conference on Robot Learning (CoRL)},
  year={2024}
}

@article{black2024pi_0,
  title={$\\pi_0$: A Vision-Language-Action Flow Model for General Robot Control},
  author={Black, Kevin and Brown, Noah and Driess, Danny and Esmail, Adnan and Equi, Michael and Finn, Chelsea and Fusai, Niccolo and Groom, Lachy and Hausman, Karol and Ichter, Brian and others},
  journal={arXiv preprint arXiv:2410.24164},
  year={2024}
}

@inproceedings{jia2025lift3d,
  title={Lift3D Policy: Lifting 2D Foundation Models for Robust 3D Robotic Manipulation},
  author={Jia, Yueru and Liu, Jiaming and Chen, Sixiang and Gu, Chenyang and Wang, Zhilve and Luo, Longzan and Li, Xiaoqi and Wang, Pengwei and Wang, Zhongyuan and Zhang, Renrui and others},
  booktitle={Proceedings of the Computer Vision and Pattern Recognition Conference (CVPR)},
  pages={17347--17358},
  year={2025}
}

@article{wan2025equivariant,
  title={Equivariant Diffusion Model With A5-Group Neurons for Joint Pose Estimation and Shape Reconstruction},
  author={Wan, Boyan and Shi, Yifei and Chen, Xiaohong and Xu, Kai},
  journal={IEEE Transactions on Pattern Analysis and Machine Intelligence},
  year={2025},
  publisher={IEEE}
}

@article{du2025se,
  title={SE (3)-Equivariance Learning for Category-level Object Pose Estimation},
  author={Du, Hongzhi and Li, Yanyan and Di, Yan and Zhang, Teng and Sun, Yanbiao and Zhu, Jigui},
  journal={IEEE Transactions on Instrumentation and Measurement},
  year={2025},
  publisher={IEEE}
}

@article{zhu2023robot,
  title={On robot grasp learning using equivariant models},
  author={Zhu, Xupeng and Wang, Dian and Su, Guanang and Biza, Ondrej and Walters, Robin and Platt, Robert},
  journal={Autonomous Robots},
  volume={47},
  number={8},
  pages={1175--1193},
  year={2023},
  publisher={Springer}
}

@inproceedings{lim2024equigraspflow,
  title={Equigraspflow: SE (3)-equivariant 6-dof grasp pose generative flows},
  author={Lim, Byeongdo and Kim, Jongmin and Kim, Jihwan and Lee, Yonghyeon and Park, Frank C},
  booktitle={Conference on Robot Learning (CoRL)},
  year={2024}
}

@inproceedings{wang2022so2,
  title = "{SO(2)-Equivariant Reinforcement Learning}",
  author = {Wang, Dian and Walters, Robin and Platt, Robert},
  booktitle = {International Conference on Learning Representations (ICLR)},
  year = {2022}
}

@inproceedings{zhao2024equivariant,
  title={Equivariant action sampling for reinforcement learning and planning},
  author={Zhao, Linfeng and Howell, Owen and Zhu, Xupeng and Park, Jung Yeon and Zhang, Zhewen and Walters, Robin and Wong, Lawson LS},
  booktitle={International Workshop on the Algorithmic Foundations of Robotics (WAFR)},
  year={2024}
}

@inproceedings{yang2024equibot,
  title={Equibot: Sim (3)-equivariant diffusion policy for generalizable and data efficient learning},
  author={Yang, Jingyun and Cao, Zi-ang and Deng, Congyue and Antonova, Rika and Song, Shuran and Bohg, Jeannette},
  booktitle={Conference on Robot Learning (CoRL)},
  year={2024}
}

@inproceedings{wang2024equivariant,
  title={Equivariant diffusion policy},
  author={Wang, Dian and Hart, Stephen and Surovik, David and Kelestemur, Tarik and Huang, Haojie and Zhao, Haibo and Yeatman, Mark and Wang, Jiuguang and Walters, Robin and Platt, Robert},
  booktitle={Conference on Robot Learning (CoRL)},
  year={2024}
}

@inproceedings{yang2024equivact,
  title={Equivact: Sim (3)-equivariant visuomotor policies beyond rigid object manipulation},
  author={Yang, Jingyun and Deng, Congyue and Wu, Jimmy and Antonova, Rika and Guibas, Leonidas and Bohg, Jeannette},
  booktitle={2024 IEEE international conference on robotics and automation (ICRA)},
  pages={9249--9255},
  year={2024},
  organization={IEEE}
}

@inproceedings{zhu2025se,
  title={SE (3)-Equivariant Diffusion Policy in Spherical Fourier Space},
  author={Zhu, Xupeng and Wang, Fan and Walters, Robin and Shi, Jane},
  booktitle={International conference on machine learning (ICML)},
  year={2025}
}

@inproceedings{ho2020denoising,
  title={Denoising diffusion probabilistic models},
  author={Ho, Jonathan and Jain, Ajay and Abbeel, Pieter},
  booktitle={Advances in neural information processing systems (NeurIPS)},
  volume={33},
  pages={6840--6851},
  year={2020}
}

@inproceedings{lipman2022flow,
  title={Flow matching for generative modeling},
  author={Lipman, Yaron and Chen, Ricky TQ and Ben-Hamu, Heli and Nickel, Maximilian and Le, Matt},
  booktitle={International Conference on Learning Representations (ICLR)},
  year={2022}
}

@inproceedings{tie2024seed,
  title={Et-seed: Efficient trajectory-level se (3) equivariant diffusion policy},
  author={Tie, Chenrui and Chen, Yue and Wu, Ruihai and Dong, Boxuan and Li, Zeyi and Gao, Chongkai and Dong, Hao},
  booktitle={International Conference on Learning Representations (ICLR)},
  year={2024}
}

@inproceedings{ryu2024diffusion,
  title={Diffusion-edfs: Bi-equivariant denoising generative modeling on se (3) for visual robotic manipulation},
  author={Ryu, Hyunwoo and Kim, Jiwoo and An, Hyunseok and Chang, Junwoo and Seo, Joohwan and Kim, Taehan and Kim, Yubin and Hwang, Chaewon and Choi, Jongeun and Horowitz, Roberto},
  booktitle={Proceedings of the Computer Vision and Pattern Recognition Conference (CVPR)},
  pages={18007--18018},
  year={2024}
}

@inproceedings{liao2022equiformer,
  title={Equiformer: Equivariant graph attention transformer for 3d atomistic graphs},
  author={Liao, Yi-Lun and Smidt, Tess},
  booktitle={International Conference on Learning Representations (ICLR)},
  year={2022}
}

@inproceedings{liao2023equiformerv2,
  title={Equiformerv2: Improved equivariant transformer for scaling to higher-degree representations},
  author={Liao, Yi-Lun and Wood, Brandon and Das, Abhishek and Smidt, Tess},
  booktitle={International Conference on Learning Representations (ICLR)},
  year={2023}
}

@inproceedings{geng2025meanflowsonestepgenerative,
  title={Mean flows for one-step generative modeling},
  author={Geng, Zhengyang and Deng, Mingyang and Bai, Xingjian and Kolter, J Zico and He, Kaiming},
  booktitle={Advances in neural information processing systems (NeurIPS)},
  year={2025}
}

@article{zhang2025alphaflowunderstandingimprovingmeanflow,
  title={AlphaFlow: Understanding and Improving MeanFlow Models},
  author={Zhang, Huijie and Siarohin, Aliaksandr and Menapace, Willi and Vasilkovsky, Michael and Tulyakov, Sergey and Qu, Qing and Skorokhodov, Ivan},
  journal={arXiv preprint arXiv:2510.20771},
  year={2025}
}

@inproceedings{prasad2024consistencypolicyacceleratedvisuomotor,
  title={Consistency policy: Accelerated visuomotor policies via consistency distillation},
  author={Prasad, Aaditya and Lin, Kevin and Wu, Jimmy and Zhou, Linqi and Bohg, Jeannette},
  booktitle={Proceedings of Robotics: Science and Systems (RSS)},
  year={2024}
}

@inproceedings{zhong2025freqpolicy,
  title={FreqPolicy: Frequency Autoregressive Visuomotor Policy with Continuous Tokens},
  author={Zhong, Yiming and Liu, Yumeng and Xiao, Chuyang and Yang, Zemin and Wang, Youzhuo and Zhu, Yufei and Shi, Ye and Sun, Yujing and Zhu, Xinge and Ma, Yuexin},
  booktitle={Advances in neural information processing systems (NeurIPS)},
  year={2025}
}

@article{yang2024consistencyflowmatchingdefining,
  title={Consistency flow matching: Defining straight flows with velocity consistency},
  author={Yang, Ling and Zhang, Zixiang and Zhang, Zhilong and Liu, Xingchao and Xu, Minkai and Zhang, Wentao and Meng, Chenlin and Ermon, Stefano and Cui, Bin},
  journal={arXiv preprint arXiv:2407.02398},
  year={2024}
}

@inproceedings{mandlekar2023mimicgen,
  title={Mimicgen: A data generation system for scalable robot learning using human demonstrations},
  author={Mandlekar, Ajay and Nasiriany, Soroush and Wen, Bowen and Akinola, Iretiayo and Narang, Yashraj and Fan, Linxi and Zhu, Yuke and Fox, Dieter},
  booktitle={Conference on robot learning (CoRL)},
  year={2023}
}

@article{mandlekar2021matters,
  title={What matters in learning from offline human demonstrations for robot manipulation},
  author={Mandlekar, Ajay and Xu, Danfei and Wong, Josiah and Nasiriany, Soroush and Wang, Chen and Kulkarni, Rohun and Fei-Fei, Li and Savarese, Silvio and Zhu, Yuke and Mart{\'\i}n-Mart{\'\i}n, Roberto},
  journal={arXiv preprint arXiv:2108.03298},
  year={2021}
}

@article{frans2024one,
  title={One step diffusion via shortcut models},
  author={Frans, Kevin and Hafner, Danijar and Levine, Sergey and Abbeel, Pieter},
  journal={arXiv preprint arXiv:2410.12557},
  year={2024}
}

@inproceedings{he2024consistency,
  title={Consistency diffusion bridge models},
  author={He, Guande and Zheng, Kaiwen and Chen, Jianfei and Bao, Fan and Zhu, Jun},
  booktitle={Advances in Neural Information Processing Systems (NeurIPS)},
  volume={37},
  pages={23516--23548},
  year={2024}
}

@inproceedings{hu2024adaflow,
  title={Adaflow: Imitation learning with variance-adaptive flow-based policies},
  author={Hu, Xixi and Liu, Qiang and Liu, Xingchao and Liu, Bo},
  booktitle={Advances in Neural Information Processing Systems (NeurIPS)},
  volume={37},
  pages={138836--138858},
  year={2024}
}

@inproceedings{klein2023equivariant,
  title={Equivariant flow matching},
  author={Klein, Leon and Kr{\"a}mer, Andreas and No{\'e}, Frank},
  booktitle={Advances in Neural Information Processing Systems (NeurIPS)},
  volume={36},
  pages={59886--59910},
  year={2023}
}

@inproceedings{xu2026hero,
  title={HeRO: Hierarchical 3D Semantic Representation for Pose-aware Object Manipulation},
  author={Xu, Chongyang and Cheng, Shen and Li, Haipeng and Fan, Haoqiang and Feng, Ziliang and Liu, Shuaicheng},
  booktitle={2024 IEEE international conference on robotics and automation (ICRA)},
  year={2026} 
}

@article{xu2026actiongeometryprediction3dgeometric,
      title={Action-Geometry Prediction with 3D Geometric Prior for Bimanual Manipulation}, 
      author={Chongyang Xu and Haipeng Li and Shen Cheng and Jingyu Hu and Haoqiang Fan and Ziliang Feng and Shuaicheng Liu},
      year={2026},
      journal={arXiv preprint arXiv:2410.12557},
}

@inproceedings{RAW-Flow,
    title={RAW-Flow: Advancing RGB-to-RAW Image Reconstruction with Deterministic Latent Flow Matching},
    author={Liu, Zhen and Feng, Diedong and Jiang, Hai and Zeng, Liaoyuan and Wang, Hao and Feng, Chaoyu and Lei, Lei and Zeng, Bing and Liu, Shuaicheng},
    booktitle={Proceedings of the AAAI Conference on Artificial Intelligence (AAAI)},
    volume={40},
    number={9},
    pages={7431-7439},
    year={2026},
}

@inproceedings{du2025hierarchical,
  title={Hierarchical neural semantic representation for 3d semantic correspondence},
  author={Du, Keyu and Hu, Jingyu and Li, Haipeng and Xu, Hao and Huang, Haibin and Fu, Chi-Wing and Liu, Shuaicheng},
  booktitle={Proceedings of the SIGGRAPH Asia 2025 Conference Papers},
  pages={1--11},
  year={2025}
}

@article{luo2026DMAligner,
  title={DMAligner: Enhancing Image Alignment via Diffusion Model Based View Synthesis},
  author={Luo, Xinglong and Luo, Ao and Wang, Zhengning and Yang, Yueqi and Feng, Chaoyu and Lei, Lei and Zeng, Bing and Liu, Shuaicheng},
  journal={arXiv preprint arXiv:2602.23022},
  year={2026}
}
}

\clearpage
\setcounter{page}{1}
\maketitlesupplementary


\appendix
\renewcommand{\thesection}{\Alph{section}}

\renewcommand{\thefigure}{\Alph{section}.\arabic{figure}}
\setcounter{figure}{0}

\renewcommand{\thetable}{\Alph{section}.\arabic{table}}
\setcounter{table}{0}

\section{Real-World Robot Experiments Details}
\label{sec:rationale}
\begin{figure}[t]
    \centering
    \includegraphics[width=\linewidth]{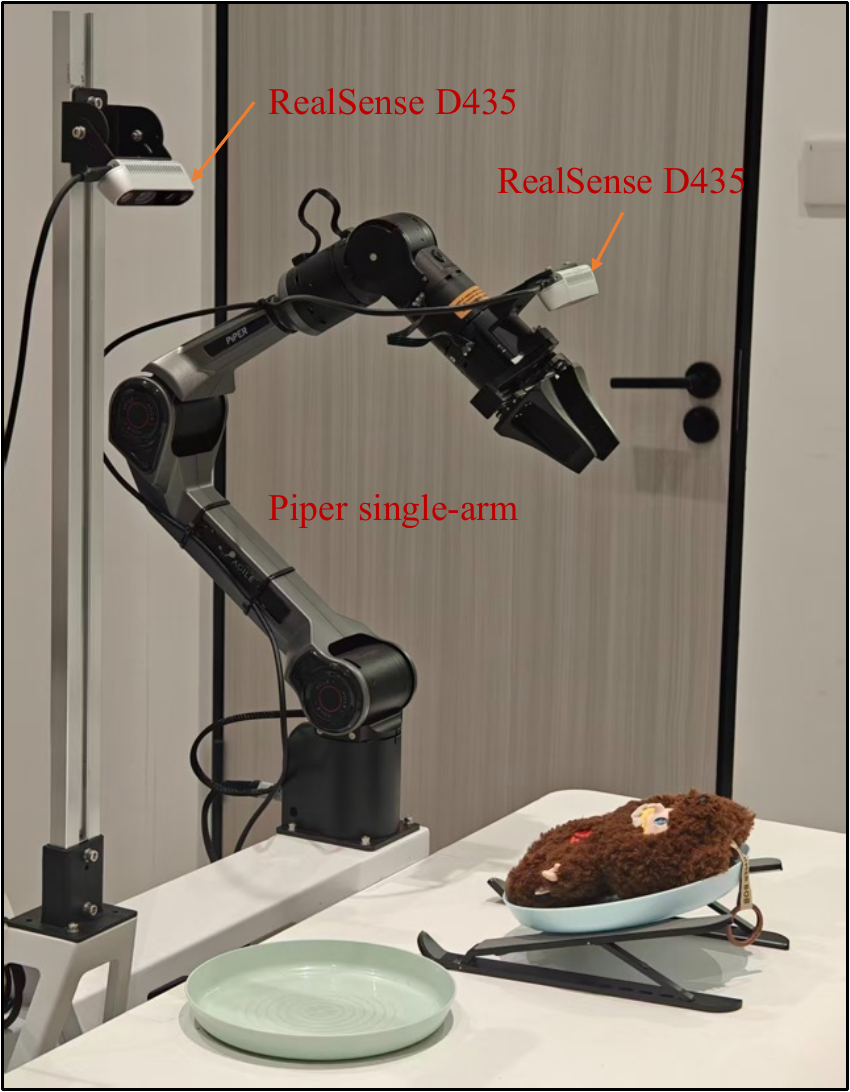}
    \caption{The overall setup of the real-world robotic experiment platform includes two RealSense D435 RGB-D cameras, a 6-DOF PIPER robotic arm equipped with a two-finger gripper, and a tabletop workspace.}
    \label{realrobot}
\end{figure}

\paragraph{Real-World Robot Experiments Details.}
We evaluate the effectiveness of E3Flow on 4 real-world physical manipulation tasks: Storing Toy, Bottle Place, Stack Blocks, and Assembly. The details of the tasks are described in Sec.~\ref{sec:task Details}.
Fig.~\ref{realrobot} shows our real-world robotic experimental setup, which consists of two RealSense D435 RGB-D cameras (a head camera and a hand camera), a 6-DOF PIPER robotic arm equipped with a two-finger gripper, and a white tabletop workspace. For each task, we collected 50 expert demonstrations using a teleoperation (puppet-master) setup, recording 3D information from the head camera and RGB image observations from the hand camera. The image observations from the hand camera have a resolution of 480×640, and the head camera point cloud contains 1024 points after cropping. We trained SDP \cite{zhu2025se}, EquiDiff \cite{wang2024equivariant}, and DP \cite{chi2025diffusion} as baseline methods for comparison with E3Flow, and evaluated each task over 20 rollouts. Table~\ref{tab:realsr} reports the success rates for all tasks. The results show that E3Flow achieves the highest average success rate. 

\paragraph{Results Analysis.}
We further analyzed the failure modes of each method across tasks. DP lacks orientation-awareness, resulting in poor performance on tasks that are highly sensitive to object orientation. EquiDiff achieves relatively high success on the Assembly task, but still exhibits limitations when handling tasks with significant SE(3) variation. The primary failure mode of SDP is inaccurate object localization, which leads to failed grasps or incorrect placements, an issue largely attributed to occlusions present in single-view point clouds. In contrast, E3Flow effectively mitigates point cloud occlusions by leveraging heterogeneous visual modalities, outperforming the strongest baseline, SDP. Moreover, compared with EquiDiff and DP, the continuous SE(3) representation in E3Flow demonstrates stronger robustness in real-world environments. Overall, the real-world experiments further validate the applicability and effectiveness of E3Flow.

\section{Task Details}
\label{sec:task Details}
\begin{figure*}[t]
    \centering
    \includegraphics[width=\linewidth]{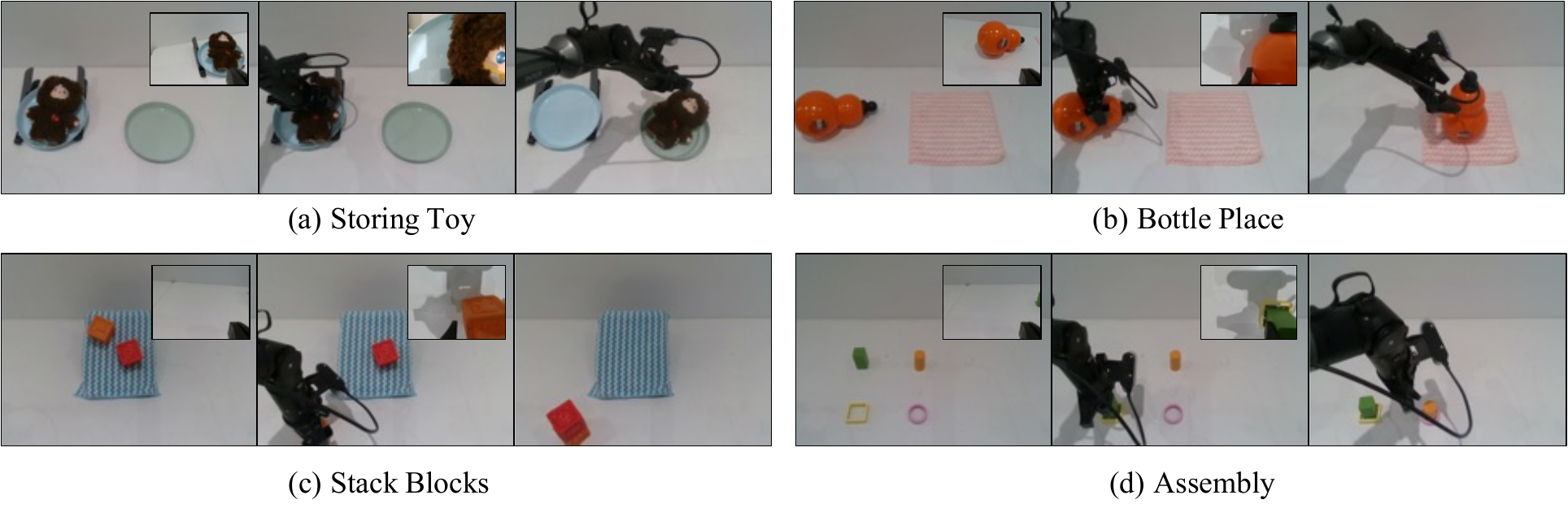}
    \caption{Visualization of the execution process for the real-world robotic tasks.}
    \label{realtask}
\end{figure*}

The visualization of the real-world robot task execution process are illustrated in Fig.~\ref{realtask}.
\paragraph{Storing Toys.} A toy is placed on an inclined plane, and the robot must pick it up from the slope and place it into a bowl on the flat tabletop. The incline angle varies between 15° and 60°, and the in-plane positions of both the toy and the slope are randomly initialized within a 10 cm range. In addition, the toy is randomly rotated around the slope surface to simulate SE(3) variations. Successfully completing the Storing Toys task requires the robot to extract rich orientation information.

\begin{table*}[th]
\centering
\caption{Success rates of different methods on various tasks.}
\label{tab:realsr}
\begin{tabular}{lcccccc}
\toprule
Method & Storing Toys & Bottle Place & Stack Blocks & Assembly & Average SR (\%)\\
\midrule
E3Flow (ours)                       & 70 & 80 & 95 & 60 & 76  \\
SDP \cite{zhu2025se}                & 60 & 65 & 85 & 40 & 62  \\
EquiDiff \cite{wang2024equivariant} & 5 & 25 & 45 & 45  & 30   \\
DP \cite{chi2025diffusion}          & 0 & 10 & 35 & 20  & 16   \\
\bottomrule
\end{tabular}
\end{table*}

\paragraph{Bottle Place.} An irregular-shaped bottle is placed on the tabletop with a random SE(3) orientation. The robot must pick up the bottle and place it upright on a mat, whose position is randomly initialized within a 10 cm range on the tabletop. Placing the bottle upside down on the mat is not counted as a success.

\paragraph{Stack Blocks.} Two blocks of different colors are randomly placed on a blue inclined plane, whose tilt angle varies between 15° and 60°. The robot must first pick up one block and place it on the tabletop, then pick up the second block and stack it on top of the first block to complete the task. The initial positions of the blocks are randomly sampled within the inclined plane, and the stacking location can be anywhere on the tabletop.

\paragraph{Assembly.} A cylinder and a cuboid are placed on the tabletop, along with a circular slot and a square slot that correspond to them, respectively. The robot must insert the cylinder into the circular slot and place the cuboid into the square slot, simulating a two-step assembly task. The cylinder and cuboid are randomly positioned on the left and right sides of the tabletop, and the task is considered successful only when both assembly actions are completed.

\section{More Implementation Details}
\label{sec:Details}

\begin{table*}[t]
\centering
\caption{Hyperparameter settings for different methods.}
\label{tab:hyperparameters}
\begin{tabular}{lccccccc}
\toprule
 & \textbf{E3Flow} & \textbf{SDP(DDPM)} & \textbf{SDP(DDIM)} & \textbf{EquiDiff} & \textbf{EquiBot} & \textbf{DP} & \textbf{DP3} \\
\midrule
Batch Size                 & 64 & 64  & 64  & 128  & 128  & 128 & 128 \\
Image Size                 & 84 & -  & -  & -  & 84  & - & - \\
Point Clouds                 & 1024 & 1024  & 1024  & 1024  & 1024  & 1024 & 1024 \\
Prediction Horizon         & 16 & 16  & 16  & 16  & 16  & 16  & 16  \\
Action Horizon             & 8 & 8   & 8   & 8   & 8   & 8   & 8   \\
Learning Rate              & 1e-4 & 1e-4 & 1e-4 & 1e-4 & 1e-4 & 1e-4 & 1e-4 \\
Epochs                     & 500 & 500 & 500 & 500 & 500 & 500 & 500 \\
Learning Rate Scheduler    & cosine & cosine & cosine & cosine & cosine & cosine & cosine \\
Noise Scheduler            & - & DDPM & DDIM & DDPM & DDPM & DDPM & DDIM \\
Inference Step             & 10 & 100 & 10 & 100 & 100 & 100 & 100 \\
Visual Encoded Dimension    & 128 & 128 & 128 & 128 & 128 & 128 & 64 \\
\bottomrule
\end{tabular}
\end{table*}

\begin{table}[h]
\centering
\caption{Network implementation Details.}
\label{tab:Details}
\setlength{\abovecaptionskip}{1pt}
\setlength{\belowcaptionskip}{1pt}
\setlength{\tabcolsep}{4pt}
\begin{tabular}{l l}
\toprule
\textbf{Module} & \textbf{Details} \\
\midrule
Image Encoder & ResNet-18 \\
Point Cloud Encoder & 5-layer ResNet + EquiformerV2 \\
Point Cloud Features & 128-dim \\
FEM Output & 128-dim \\
Equi-Flow U-Net & SDTU w/ spherical Fourier conv \\
Equivariant Layers & Equivariant Linear + FiLM \\
\bottomrule
\end{tabular}
\end{table}

\paragraph{Hyperparameters} To ensure the reproducibility of our experiments, we provide additional experimental details in Table~\ref{tab:hyperparameters}, including the hyperparameter settings for each component. Most of the configurations follow those used in SDP \cite{zhu2025se}. In addition, we run three random seeds for each task. The reported success rate is the average of the highest success rates obtained across the three seeds. It is worth noting that, for each method, we perform an evaluation every 20 epochs. Each evaluation consists of 50 episodes, and each episode share identical initial scenarios to guarantee fairness and to justify the selection of the maximum success rate. This evaluation protocol is identical to that of SDP \cite{zhu2025se}.

\paragraph{Network Details} 
Table~\ref{tab:Details} presents more details of the network. For the visual encoders, we follow the configurations used in SDP \cite{zhu2025se} and EquiDiff \cite{wang2024equivariant}. Specifically, we use ResNet-18 to extract features from the eye-in-hand RGB images, and EquiformerV2 to extract equivariant point-cloud features. The feature enhancement module (FEM) is applied only to Type-0 (scalar) features, leaving Type-1 (vector) features and Type-2 (higher-order equivariant) features unchanged. 

For the Equi-Flow U-Net, we adopt the Spherical Denoising Temporal U-Net (SDTU) proposed by Zhu \cite{zhu2025se}. SDTU is a 1D U-Net constructed in the spherical Fourier domain with spatio-temporal equivariance. Temporal equivariance is achieved via 1D convolutions along the time dimension \(t\), while spatial (SO(3)) equivariance is ensured by performing channel mixing temporal convolutions independently for each spherical harmonic degree Type-\(\ell\) (i.e. independently for each irreducible representation), so that the SO(3) action does not mix different \(\ell\) subspaces. Concretely, the spherical Fourier temporal convolution can be written as
\begin{equation}
\tilde{h}^{\,o}_{\ell,m,t}
    \;=\;
    \sum_{j=0}^{R} 
    \sum_{i \in \mathcal{I}}
        \tilde{h}^{\,i}_{\ell,m,\,t-j}\,
        \omega^{\,i \rightarrow o}_{\ell,\,j},
\label{eq:fourier_conv_corrected}
\end{equation}
where \(i\) and \(o\) index input and output feature channels, and the pair \((\ell,m)\) denotes the degree and order of the spherical Fourier component \(\tilde{h}\). The set \(\mathcal{I}\) contains all input channels, and \(j\) indexes the temporal lag up to \(R\). Importantly, the learnable weights \(\omega^{\,i\rightarrow o}_{\ell,\,j}\) are independent of \(m\); this independence is required for SO(3)-equivariance. By Schur’s lemma, any linear operator that commutes with the SO(3) group action must act as a scalar multiple of the identity on each irreducible subspace. Therefore, the above convolution acting only within each \(\ell\) subspace with weights independent of \(m\) is SO(3)-equivariant.

Accordingly, we extend the feature modulation layer into an equivariant FiLM layer using equivariant linear layers, and apply conditional modulation to each feature type separately, ensuring equivariance throughout the entire flow matching process. 
\begin{align}
\mathrm{EFiLM}(h_\ell \mid \gamma_\ell,\beta_\ell) =\left(
   \gamma_\ell^\top h_\ell\;
   \frac{h_\ell}{\|h_\ell\|}
   + \beta_\ell
   \right)
\end{align}

The key property of the equivariant FiLM layer is that applying the rotation before EFiLM produces the same result as applying EFiLM before the rotation.
This can also be established using Schur’s lemma. For a group element $g$ and an $\ell$-type feature $h_\ell$, we examine the EFiLM operation under the action of the Wigner $D$-matrix:

\begin{align}
\mathrm{EFiLM}&\!\left(
D_\ell(g)h_\ell \,\middle|\,
D_\ell(g)\gamma_\ell,\; D_\ell(g)\beta_\ell
\right)
\\[2pt]
&= (D_\ell(g)\gamma_\ell)^\top D_\ell(g)h_\ell\;
   \frac{D_\ell(g)h_\ell}{\|D_\ell(g)h_\ell\|}
   + D_\ell(g)\beta_\ell 
\\[2pt]
&= \gamma_\ell^\top D_\ell(g)^\top D_\ell(g)h_\ell\;
   \frac{D_\ell(g)h_\ell}{\|D_\ell(g)h_\ell\|}
   + D_\ell(g)\beta_\ell .
\end{align}

Using\ the\ orthogonality\ of\ Wigner\ D\text{-}matrices,\
$\|D_\ell(g)h_\ell\|=\|h_\ell\|$, we\ have
\begin{align}
\mathrm{EFiLM}&\!\left(
D_\ell(g)h_\ell \,\middle|\,
D_\ell(g)\gamma_\ell,\; D_\ell(g)\beta_\ell
\right)
\\[2pt]
&= \gamma_\ell^\top h_\ell\;
   \frac{D_\ell(g)h_\ell}{\|h_\ell\|}
   + D_\ell(g)\beta_\ell 
\\[2pt]
&= D_\ell(g)\!
   \left(
   \gamma_\ell^\top h_\ell\;
   \frac{h_\ell}{\|h_\ell\|}
   + \beta_\ell
   \right)
\\[2pt]
&= D_\ell(g)\cdot
   \mathrm{EFiLM}(h_\ell \mid \gamma_\ell,\beta_\ell).
\end{align}

This final line follows directly from Schur's lemma. Overall, the above derivation demonstrates that the Equi-Flow U-Net is equivariant with respect to both its inputs and outputs, making it applicable to flow matching processes rather than being limited solely to diffusion-based approaches.

\section{End-to-End Symmetry Analysis}
We analyze the equivariant properties of E3Flow. First, the point cloud encoder is SO(3)-equivariant, extracting spherical features including Type-0 scalar features, Type-1 vector features, and Type-2 higher-order tensor features, with Type-1 and Type-2 features being equivariant. The image encoder is not equivariant and extracts visual detail features that indicate potential contact regions. Although the feature enhancement module (FEM) is not equivariant, it only injects visual detail features into type-0 features, thus preserving the equivariant part of the spherical harmonic features. Consequently, the output of the visual encoder remains equivariant. Furthermore, Sec.~\ref{sec:Details} provides a detailed analysis showing that the Equi-Flow U-Net is SO(3)-equivariant. Therefore, the vector field predicted by the network is equivariant. In summary, E3Flow is an end-to-end SE(3)-equivariant model, with translational equivariance realized through coordinate normalization.

\end{document}